\documentclass[letterpaper]{bioinfo}
\copyrightyear{2015} \pubyear{2015}

\access{Advance Access Publication Date: Day Month Year}
\appnotes{Manuscript Category}

\usepackage{array} 
\usepackage{url}            
\usepackage{booktabs}       
\usepackage{amsfonts}       
\usepackage{nicefrac}       
\usepackage{microtype}      
\usepackage{bm}             
\usepackage{color}
\usepackage[pdf]{graphviz}
\usepackage{wrapfig}		
\usepackage{float}
\usepackage{fullpage}

\usepackage{microtype}
\usepackage{tabularx, booktabs} 
\usepackage{caption}

\newcolumntype{D}{ >{\centering\arraybackslash} m{0.2cm} }

\begin{document}

\subtitle{}

\title[Dropout Feature Ranking]{Dropout Feature Ranking for Deep Learning Models}
\author[Chang \textit{et~al}.]{Chun-Hao Chang
\,$^{\text{\sfb 1,2,3}*}$
, Ladislav Rampasek\, $^{\text{\sfb 1,2,3}}$ and Anna Goldenberg\,$^{\text{\sfb 1,2,3,}*}$}
\address{$^{\text{\sf 1}}$Computer Science, University of Toronto, Toronto, Canada, \\
$^{\text{\sf 2}}$Vector Institute, Toronto, Canada, \\
$^{\text{\sf 3}}$Sickkids Hospital, Toronto, Canada.}

\corresp{$^\ast$To whom correspondence should be addressed.}

\history{}

\editor{}

\abstract{
Deep neural networks (DNNs) achieve state-of-the-art results in a variety of domains. 
Unfortunately, DNNs are notorious for their non-interpretability, and thus limit their applicability in hypothesis-driven domains such as biology and healthcare.
Moreover, in the resource-constraint setting, it is critical to design tests relying on fewer more informative features leading to high accuracy performance within reasonable budget.
We aim to close this gap by proposing a new general feature ranking method for deep learning. 
We show that our simple yet effective method performs on par or compares favorably to eight strawman, classical and deep-learning feature ranking methods in two simulations and five very different datasets on tasks ranging from classification to regression, in both static and time series scenarios. 
We also illustrate the use of our method on a drug response dataset and show that it identifies genes relevant to the drug-response.
}

\maketitle
\section{Introduction}

Deep neural networks (DNNs) have started to come out as top performers in biology and healthcare including genomics \citep{2015_brendon_frey}, medical imaging \citep{2017_sebastian}, EEG \citep{2017_andrew_ng} and EHR \citep{2017_duke_gp_rnn}. 
However, DNNs are black-box models and notorious for their non-interpretability. 
In the fields of biology and healthcare, to derive hypotheses that could be experimentally verified, it is paramount to provide information about which biological or clinical features are driving the prediction. 
The desired data may be very expensive to collect, thus it is also important to generate experimental designs that will collect the most effective data leading to the highest accuracy within reasonable budget. 
Therefore, there is a strong need for feature ranking for deep learning methods to advance their use in biology and healthcare. We aim to close this gap by proposing a new general feature ranking method for deep learning.


In this work we propose to rank features by variational dropout \citep{gal_concrete_2017}. 
Dropout is an effective technique commonly used to regularize neural networks by randomly removing a subset of hidden node values and setting them to $0$. 
In this work we use the Dropout concept on the input feature layer and optimize the corresponding feature-wise dropout rate. 
Since each feature is removed stochastically, our method creates a similar effect to feature bagging \citep{1995_random_bagging} and manages to rank correlated features better than other non-bagging methods such as LASSO. 
We compare our method to Random Forest (RF), LASSO, ElasticNet, Marginal ranking and several techniques to derive importance in DNN such as Deep Feature Selection and various heuristics. 
We first test it on $2$ simulation datasets and shows that our methods can rank features correctly in the non-linear feature interactions especially among the important features. 
Then we test it on $4$ real-world datasets and show that our method has higher performance under the same number of features in the deep neural network.
Then we test it on a multivariate clinical time-series dataset and show that our method also rivals or outperforms other methods in recurrent neural network setting.
Finally, we test our method on a real-world drug response prediction problem using a previously proposed Variational Autoencoder (VAE) \citep{kingma2013auto}. 
In this proof-of-concept application, we show that our method identifies genes relevant to the drug-response.

\section{Related Work}
Many previously proposed approaches to interpret DNNs focus on interpreting a decision (such as assigning a particular classification label in an image) for a specific example at hand (e.g. \citep{simonyan2013deep, zeiler2014visualizing,  ribeiro_why_2016, zhou2016learning, selvaraju2016grad,  2017_deeplift, zintgraf2017visualizing, 2017_bbmp, dabkowski2017real}). 
In this case, a method would aim to figure out which parts of a given image make the classifier think that this particular image should be classified as a dog. 
These methods are unfortunately not easy to use for the purpose of feature selection or ranking, where the importance of the feature should be gleaned across the whole dataset.


Several works have mentioned using variational dropout to achieve better performance \citep{gal_concrete_2017} \cite{kingma2015variational}, have a Bayes interpretation of dropout \citep{maeda_bayesian_2014}, or compress the model architecture \citep{molchanov2017variational}. These works focus on tuning the dropout rate to automatically get the best performance, but do not consider applying it to the feature ranking problems.

\citet{2015_wasserman_dfs} proposed Deep Feature Selection (Deep FS). Deep FS adds another hidden layer to the network with one connection per input node to this hidden layer (of the same size as input) and uses an $\ell_1$ penalty on this layer. The weights between these layers are initialized to $1$ but since they are not constrained to $[0,1]$, they can become large positive and negative values. Thus, this additional layer can amplify a particular input and will need to be balanced within the original network architecture. Additionally, using $\ell_1$ penalty prevents Deep FS from selecting correlated features, important in many biological and health applications.

Finally, several works also targeted interpreting features in a clinical setting. \citet{che2015distilling} uses Gradient-Boosted Trees to mimic a recurrent neural network on a healthcare dataset to achieve comparable performance. \citet{2016_ae_dfs} interprets the clinical features by autoencoder and random forest. \citet{suresh2017marzykeh} use Recurrent Neural Network to predict the clinical dataset and use the ranking heuristics called 'Mean' in our settings. These approaches rely on additional decision tree architecture to learn the features, or use heuristics which have a weaker ranking performance in our experiments. 

\section{Methods}
\subsection{Variational Dropout}
Dropout \citep{srivastava2014dropout} is one of the most effective and widely used regularization techniques for neural networks. The mechanism is to inject a multiplicative Bernoulli noise for each hidden unit within a neural network. Specifically, during forward pass, for each hidden unit $k$ in layer $j$ a dropout mask $z_{jk} \sim \mathrm{Bern}(z|\theta_{jk})$ is sampled. The original hidden node value $h_{jk}$ is then multiplied by this mask $h'_{jk} = h_{jk}z_{jk}$, which stochastically sets the hidden node value to $h_{jk}$ or $0$. 

Variational dropout \citep{maeda_bayesian_2014} optimizes the dropout rate $\theta$ as a parameter instead of it being a fixed hyperparameter. 
For a neural network $f(\bm{x})$, given a mini-batch of size $M$ (sampled from training set of $N$ samples) and a dropout mask $\bm{z}$, the loss objective function that follows from the variational interpretation of dropout can be written as:
\begin{equation}
L(\theta) = - \frac{1}{M} \sum_{i=1}^M \log p(\bm{y}_i | f (\bm{x}_i, \bm{z}_i)) + \frac{1}{N}KL(q_\theta(\bm{z}) || p(\bm{z})).
 \label{eq:objfun_kl}
\end{equation}
Here, $\bm{z}_i \sim q_\theta(\bm{z})$, where $q_\theta(\bm{z})$ is the variational mask distribution and $p(\bm{z})$ is a prior distribution.

\begin{figure}[tbp]
   \centering
   \begin{tabular}{@{}c@{\hspace{.5cm}}c@{}c@{\hspace{.5cm}}c@{}}
       \includegraphics[page=1,height=100pt,angle=90,origin=a]{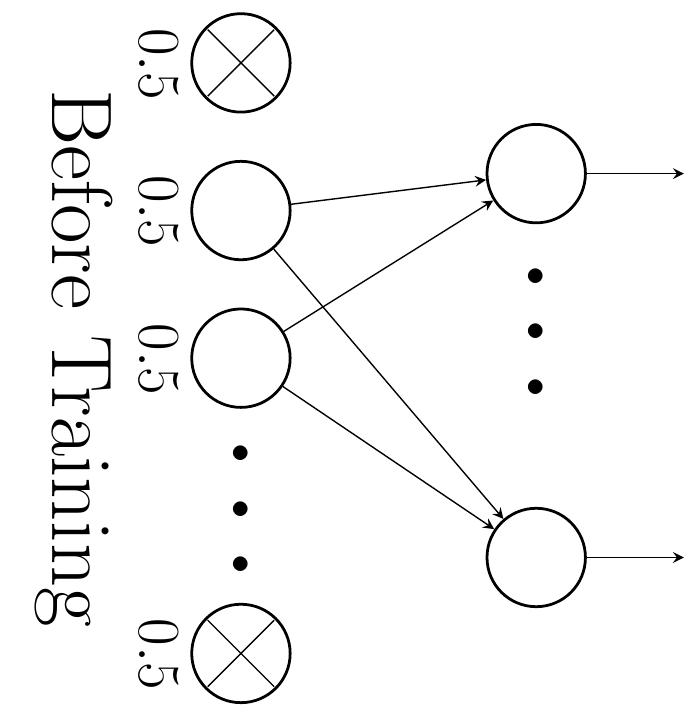} & 
       \includegraphics[page=1,height=100pt,angle=90,origin=a]{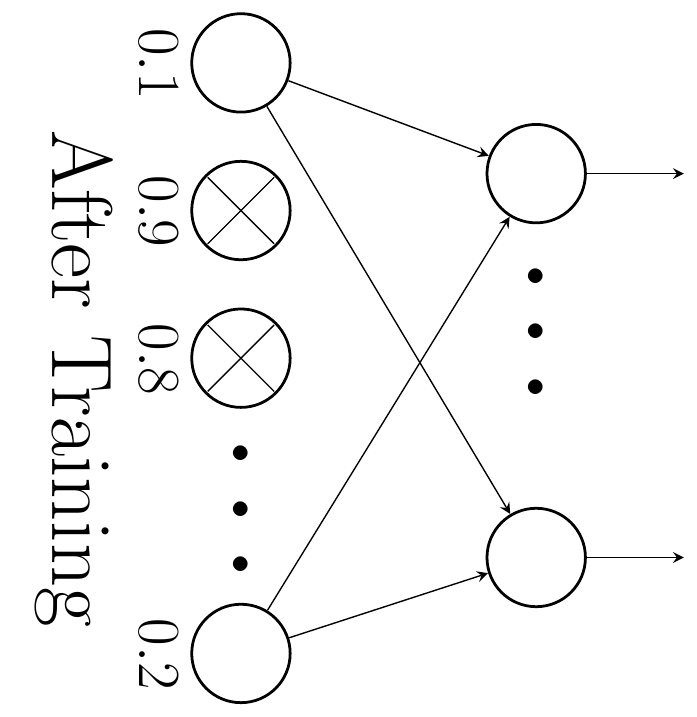} &
       \includegraphics[page=1,height=100pt,origin=a]{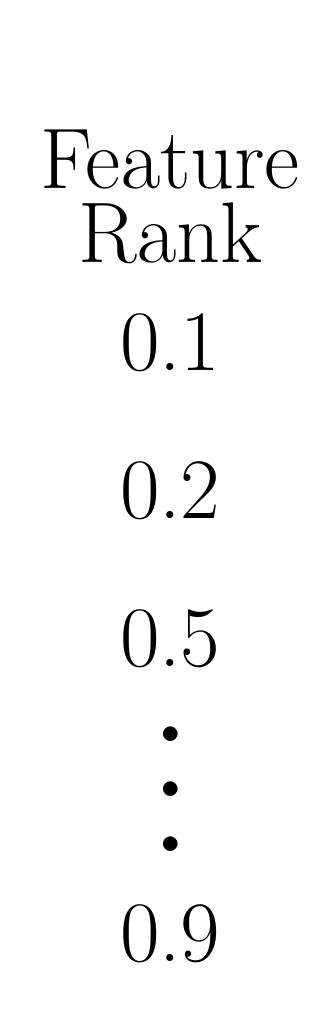} \\
   \end{tabular}
 \caption{Dropout feature ranking diagram. Before training (Left), the dropout rate for each feature is initialized to $0.5$. After training (Right), each feature gets a different dropout rate. We then rank all features based on the magnitude of the dropout rate - the lower the magnitude, the higher the rank.}
 \label{fig:method_diagram}
\end{figure}

\subsection{Feature Ranking Using Variational Dropout}
\label{sec:dfr_method}
Figure \ref{fig:method_diagram} shows our approach.
To analyze which features are important for a given pre-trained model $\mathcal{M}$ to correctly predict its target variable $y$, we introduce Dropout Feature Ranking (Dropout FR) method. In our method we add variational dropout regularization to the input layer of $\mathcal{M}$.
To achieve minimum loss, the Dropout FR model should learn small dropout rate for features that are important for correct target prediction by the analyzed model $\mathcal{M}$, while increasing the dropout rate for the rest of unimportant features. Specifically, given $D$ features,  we set a variational mask distribution $q_{\bm{\theta}}(\bm{z}) = \prod_{j=1}^D q(z_j | \theta_j) = \prod_{j=1}^D \mathrm{Bern}(z_j | \theta_j)$ as a fully factorized distribution. This gives us a feature-wise dropout rate $\theta_j$ where magnitude indicates the importance of feature $j$.

Instead of having a $KL(q_\theta(\bm{z}) || p(\bm{z}))$ in the equation \ref{eq:objfun_kl} to regularize the dropout distribution $q_\theta(\bm{z})$, we directly penalize the number of existing features (features not dropped-out). This avoids the need to set the prior dropout rate $p(\bm{z})$ and is aligned with the $\ell_0$ penalty for linear regression \citep{2012_murphy}.
Our loss function can thus be written as:
\begin{equation}
 L(\theta) = - \frac{1}{M} \sum_{i=1}^M \log p_{\mathcal{M}}(\bm{y_i} | f (\bm{x}_i \odot \bm{z_i})) + \frac{\lambda}{M} \sum_{i=1}^M \sum_{j=1}^D z_{ij}  
 \label{eq:objfun}
\end{equation}
where $\bm{z}_{i} \sim q_{\theta}(\bm{z})$ and $\lambda$ is determined by cross validation. 

\paragraph{Concrete relaxation} To optimize $L(\theta)$ w.r.t. the parameters $\theta$, we need to backpropagate through discrete variable $\bm{z}$.
We adopt the same approach as in \citet{gal_concrete_2017} to optimize our dropout rate. Specifically, instead of sampling the discrete Bernoulli variable, we sample from Concrete distribution \citep{jang2016categorical, 2016_concrete} with some temperature $t$ which produces value between $0$ and $1$.
This distribution places most of the mass in $0$ and $1$ to approximate the discrete distribution. The concrete relaxation $\tilde{z}$ for Bernoulli distribution $\mathrm{Bern}(z | \theta)$ is:
$$ 
\tilde{z} = \mathrm{sigmoid}(\frac{1}{t} (\log \theta - \log (1 - \theta) + \log u - \log (1 - u) ))
$$
where $u \sim \mathrm{Uniform}(0, 1)$. We fix $t$ as $0.1$ and find it works well in all our experiments.
Compared to traditional REINFORCE estimator \citep{1992_reinforce}, this concrete distribution has lower variance and has better performance (data not shown), so we apply it in all our experiments.

\paragraph{Annealing}
We adopt the annealing tricks to avoid the model being penalized heavy before it fully optimizes. Specifically, we increase $\lambda$ linearly from $0$ to its specified value in the first few epochs during optimization. This is similar to the KL annealing tricks \citep{bowman2015generating} in the VAE.

\paragraph{Relation to reinforcement learning}
Our method can be seen as a policy gradient based method \citep{sutton2000policy} (one of the reinforcement learning techniques) applied to feature selection setting. From this perspective, our policy is the factorized Bernoulli distribution, and the reward consists of the log probability of targets and the number of features used.
We optimize the policy that outputs the best feature combination in this large feature spaces with $2^P$ combinations where $P$ is the total number of features. 
To get the feature-wise explanation, we adopt the factorized Bernoulli distribution to gain per-feature importance value as our ranking.


\subsection{Training details of neural networks}
\label{sec:nn_details}
We list all the hyperparameters of our experiments (No Interaction, Interaction, Support2, MiniBooNE, Online News, YearPredictionMSD,  and Physionet) in table \ref{table:nn_hyperparam}. For all the feed-forward neural network, we add dropout and batch normalization in every hidden layer, and use learning rate decay and early stopping to train the classifier. For recurrent neural network, we apply dropout and batch normalization in the output. We do the grid search and cross validation to select the $\lambda$. We add a small $\ell_2$ penalty to reduce overfitting. We use Adam \citep{kingma2014adam} to optimize all the experiments. All the hyperparameters are selected by hands without much tuning.

\begin{table*}[tbp]
  \centering
  \caption{Neural Network hyperparameters for each dataset}
  \label{table:nn_hyperparam}
  \begin{tabular}{|>{\em}l|l|l|l|l|l|l|l|}
    \toprule
    \cmidrule{1-2}
 & No Interaction & Interaction & Support2 & MiniBooNE & Online News & YearPredictionMSD & Physionet \\
    \midrule
Architecture & MLP & MLP & MLP & MLP & MLP & MLP & GRU \\
Loss & MSE & MSE & CE loss & CE loss & MSE & MSE & CE loss \\
Input layer & 40 & 40 & 42 & 50 & 59 & 90 & 37 \\
Hidden layers & 40-20 & 40-20 & 170-170 & 150-100-50-25 & 80-80-80 & 200-100-80-40-20 & 64 \\
\cline{2-7}
Dropout & \multicolumn{6}{c|}{Every hidden layer with 0.5 dropout rate}& input(0.3) output(0.5) \\
\cline{2-7}
BatchNorm & \multicolumn{6}{c|}{After every hidden layer}& output \\
\cline{2-7}
Learning rate & 0.001 & 0.001 & 0.001 & 0.001 & 0.001 & 5.00E-04 & 1.00E-03 \\
L2 penalty & 1.00E-05 & 1.00E-05 & 1.00E-04 & 1.00E-04 & 1.00E-05 & 1.00E-05 & 1.00E-06 \\
Patience & 3 & 3 & 5.00E+00 & 2 & 4.00E+00 & 2 & 2 \\
Lookahead & 10 & 10 & 12 & 6 & 9 & 5 & 5 \\
Epochs & 100 & 100 & 600 & 100 & 100 & 100 & 100 \\
\toprule
\multicolumn{6}{|l|}{Dropout feature ranking parameters}                   \\
\midrule
Annealing & 30 & 30 & 30 & 30 & 30 & 30 & 30 \\
$\lambda$ & 0.1 & 1 & 0.01 & 0.1 & 0.01 & 1 & 0.001 \\
learning rate & 0.001 & 0.001 & 0.002 & 0.002 & 0.002 & 0.002 & 0.01 \\
Epochs & 200 & 200 & 400 & 150 & 200 & 100 & 300 \\
    \bottomrule
  \end{tabular}
\end{table*}

\section{Results}
First, we test and compare our droupout feature ranking method (Dropout FR) in $2$ simulation settings.
Second, we test $2$ classification and $2$ regression real-world datasets using the feed-forward neural network. 
We then compare our approach using a clinical time series dataset by interpreting a recurrent neural network.
Finally, we apply our approach to a drug-response task to understand which genes contribute to drug response in a variational autoencoder (VAE) prediction model.

\begin{table*}[tb]
  \caption{Comparisons of different ranking methods in $2$ simulations (No Interaction, Interaction).}
  \label{table:simulation}
  \begin{center}
  \begin{sc}
  \begin{tabular}{|c|c|c|c|c|c|c|}
    \toprule
     & \multicolumn{3}{c|}{No Interactiton} & \multicolumn{3}{c|}{Interactiton} \\
    \cmidrule{1-7}
Criteria & Top 40 & Top 20 & Top 5 & Top 40 & Top 20 & Top 5 \\
    \midrule
Dropout FR & $0.933\pm0.004$ & $0.998\pm0.002$ & $0.920\pm0.084$ & $0.909\pm0.016$ & $0.990\pm0.004$ & $0.949\pm0.000$ \\
Mean & $0.928\pm0.012$ & $0.998\pm0.001$ & $0.860\pm0.055$ & $0.910\pm0.016$ & $0.990\pm0.002$ & $0.949\pm0.000$ \\
Shuffle & $0.931\pm0.006$ & $0.997\pm0.001$ & $0.820\pm0.045$ & $0.900\pm0.021$ & $0.988\pm0.003$ & $0.949\pm0.000$ \\
Deep FS & $0.902\pm0.015$ & $0.979\pm0.012$ & $0.780\pm0.130$ & $0.892\pm0.022$ & $0.958\pm0.005$ & $0.453\pm0.265$ \\
RF & $0.921\pm0.016$ & $0.996\pm0.002$ & $0.860\pm0.055$ & $0.904\pm0.013$ & $0.983\pm0.012$ & $0.885\pm0.141$ \\
Elastic Net & $0.229\pm0.275$ & $0.997\pm0.001$ & $0.860\pm0.055$ & $-0.073\pm0.138$ & $-0.130\pm0.142$ & $0.190\pm0.387$ \\
LASSO & $0.101\pm0.136$ & $0.997\pm0.001$ & $0.860\pm0.055$ & $-0.073\pm0.138$ & $-0.129\pm0.142$ & $0.190\pm0.387$ \\
Marginal & $0.902\pm0.014$ & $0.995\pm0.001$ & $0.860\pm0.055$ & $-0.041\pm0.101$ & $-0.100\pm0.131$ & $0.105\pm0.376$ \\
    \bottomrule
  \end{tabular} %
  \end{sc}
  \end{center}
  \vskip -0.1in
\end{table*}

\subsection{Compared methods}
\label{sec:other_methods}
We compare our approach to a variety of strawman methods as well as approaches commonly used for feature ranking.
LASSO uses an $\ell_1$ penalty while Elastic Net uses a mix of $\ell_1$ and $\ell_2$ penalty (we pick $\alpha=0.5$ to balance between $\ell_1$ and $\ell_2$), in which the feature importance is derived from the order each feature goes to 0 as the penalty increases. Random Forest derives its feature ranking by the average  decrease of impurity across different trees (we use $200$ trees for all experiments). Marginal ranking refers to the univariate feature analysis that ignores the interaction between features. We use t-test probability as the ranking criteria in the binary classification task, and use Pearson correlation coefficient in the regression task.
Random ranking means that we randomly assign ranks to different features, serving as a baseline in the real-world dataset evaluation in the section \ref{sec:eval_in_the_real_world_dataset} and \ref{sec:hospital}.

Deep FS \citep{2015_wasserman_dfs} was proposed specifically for interpreting deep learning models. It adds another hidden layer to the network with one connection per input node to this hidden layer (of the same size as input) and uses an $\ell_1$ penalty on this layer.
After the optimization, the magnitude of the connection weight is used as a proxy of the importance of each variable. Note that to correctly evaluate importance of each feature and to ultimately rank features, in theory the method should examine the order with which weights drop to $0$ as the $\ell_1$ penalty increases. However, this would require hundreds of manual settings of the $\ell_1$ penalty hyperparameter, which is not scalable, so we follow the authors and use the connection weight instead.
We pick the $\ell_1$ coefficient $\lambda$ also by cross validation.

Finally, we use two heuristics to rank features in a DNN. We call the first approach `Mean' method: we replace one feature at a time with the mean of the feature and rank feature importance based on the corresponding increase in the training loss. Our second method is called `Shuffle': for each feature we permute its values across the samples and evaluate importance by the increase of the training loss.

\subsection{Simulation}
We simulate two datasets to show that multi-layer neural network can capture the non-linear interactions.
First, we simulate a dataset without any feature interactions (called \textbf{'No Interaction'}).
We sample $40$ features $x_{d\in\{1,\ldots,40\}} \sim N(0,1)$, while only top $20$ features are informative of target $y$. These top $20$ features have decreasing importance with increasing Bernoulli noise $z_{d\in\{1,\ldots,20\}} \sim Bern(1. - 0.05 \cdot (d - 1))$ that stochastically sets each feature to $0$. We set our target $y = \sum_{d=1}^{20} x_d \dot z_d$. 
Thus, among the informative features, the most important feature is the 1-st and the least important feature is the 20-th. And the ground truth ranking is decreasing from 1st feature to 20-th feature with the noisy features (21th - 40th) as the least important.
We then calculate the Spearman coefficient to the ground truth as our performance metric.

To compare the effect of second order interaction, we simulate another dataset with second order feature interactions (called \textbf{'Interaction'}). Namely, we use the product of feature pairs instead of each individual feature to affect the target $y$. Specifically, we set the target $y = \sum_{d=1,3 \ldots 19} x_d \cdot x_{(d+1)} \cdot z_d$, where target $y$ depends on the product of feature pair $x_d$ and $x_{(d+1)}$.
Thus, among the informative features, the most important feature pair is 1-st and 2-nd features, with decreasing importance of 19-th and 20-th feature pair. The least important features are still the noisy features (21th - 40th).

\paragraph{Training Details} We simulate $10,000$ samples for both datasets to generate enough samples for the neural network to perform reasonably well. We train a feed-forward neural network with $2$ hidden layers (exact architecture shown in section \ref{sec:nn_details}), then rank features by our Dropout FR, Mean, Shuffle and Deep FS. We train random forest with $200$ trees since we find increasing number of trees does not improve. We use 5-fold cross validation for all our experiments.

\paragraph{Result} 
In Table \ref{table:simulation}, we show the Spearman coefficients for these two datasets when comparing full features (Top $40$), only the informative features (Top $20$) and the top $5$ most informative features (Top $5$).
In the No Interaction dataset, we find all methods perform great when comparing full features except Elastic Net and LASSO. However, we find these $2$ methods perform well when only considering the top $20$ informative features. It shows that these methods can not distinguish noisy from true features, but are able to rank the strengths of informative features.
In the Interaction dataset, we find Elastic Net, LASSO and Marginal method perform much worse, showing these simple linear layer and single-feature statistical tests can not capture second order interaction effects.
We find deep learning based methods (Dropout FR, Mean, Shuffle) except Deep FS perform the best across all the settings.
Random Forest (RF) is able to distinguish noisy features from important features (see Top 40). However, we find it performs much worse when only considering the top $5$ features, showing that it can not correctly rank the very top and most important features, failing to capture complicated feature interactions.

\begin{table*}[t]
  \begin{center}
  \caption{UCI dataset classifier performance}
  \resizebox{2.1\columnwidth}{!}{%
  \begin{tabular}{|c|cc|l|l|l|l|}
    \toprule
    \cmidrule{1-7}
Datasets & Samples &\quad Features & NN & RF & Elastic Net & LASSO \\
    \midrule
Support2 (accuracy) & $9,105$ &\quad $47$ & $0.786\pm0.024$ & $\bm{0.830}\pm\bm{0.018}$ & $0.786\pm0.029$ & $0.785\pm0.029$ \\
MiniBooNE (accuracy) & $130,065$ &\quad $50$ & $\bm{0.943}\pm\bm{0.008}$ & $0.942\pm0.012$ & $0.907\pm0.001$ & $0.907\pm0.002$ \\
OnlineNewsPopularity (MSE) & $39,797$ &\quad $61$ & $\bm{0.873}\pm\bm{0.096}$ & $0.882\pm0.076$ & $0.925\pm0.141$ & $0.925\pm0.141$ \\
YearPredictionMSD (MSE) & $515,345$ &\quad 90 & $\bm{76.10}\pm\bm{1.44}$ & $85.67\pm1.75$ & $91.32\pm1.89$ & $91.32\pm1.89$ \\
    \bottomrule
  \end{tabular} %
  }
  \end{center}
  \label{table:uci_performance2}
\end{table*}

\subsection{Evaluation in the Real-World datasets}
\label{sec:eval_in_the_real_world_dataset}

\paragraph{Evaluation Criteria}
To understand which feature ranking is the best in the real-world datasets, we evaluate each feature ranking by its test set performance of top $N$ features.
A better feature ranking should reach higher performance by using same amounts of features.
We evaluate the feature ranking with two settings. 
We call the first setting \textbf{'zero-out'}: after taking top $N$ features, we set the rest of the features to $0$ and evaluate the test performance using the already trained neural network:
this represents how well we interpret a given classifier.
The second setting is called \textbf{'retrain'}: we retrain neural network using the top $N$ features.
It represents in general which features are important under this neural network architecture.

In this experiment, we evaluate our ranking approaches on 2 classification tasks (clinical dataset Support2\footnote{\url{http://biostat.mc.vanderbilt.edu/wiki/Main/DataSets}}, UCI MiniBooNE datasets), and 2 UCI regression tasks (Online News Popularity \citep{fernandes2015proactive}, YearPredictionMSD). 
Here we describe each dataset.
Support2 is a multivariate clinical dataset which aims to predict in-hospital death by patient's demographics, clinical assessments and lab tests. It consists of $9,105$ samples and $47$ features with $2,360$ positive mortality labels. MiniBooNE aims to predict effective particles in distinguish electron neutrinos (signal) from muon neutrinos (background). It consists of $130,065$ samples and $50$ features with $36,499$ positive labels.
Online News Popularity dataset predicts the sharing times of articles in the website Mashable by article topics, word compositions and timestamps. It consists of $39,797$ samples and $61$ features, and the goal is to predict the number of times this article is shared.
YearPredictitonMSD is a task to predict the published year of songs that is published from 19 to 20 century by various sound features. It consists of $515,345$ samples and $90$ song features.


We preprocess all the continuous features by clipping the values of outliers to the outlier threshold defined by the interquantile ranges (IQR) method. Then we normalize them to $0$ mean and unit variance. We also remove the outliers and normalize the target variable in Online News Popularity dataset. For categorical variable, we do the one-hot encoding.
We do 5-fold cross validation with $10$ percent training set as the validation set in all our experiments.

We show each dataset summary statistics and classifiers' performance in Table \ref{table:uci_performance2}. We select datasets that have a relatively large number of instances, the scenario where neural networks commonly outperform their competitors. With the exception of the largest dataset in this experiment (YearPredictionMSD), neural network performance is relatively close to the performance of random forest. RF even outperforms NN on the Support2, the smallest dataset. As expected, neural network performance gets better as the datasets get larger.

\begin{figure*}[tb]
   \centering
\begin{tabular}{Dc@{\hskip -2pt}c@{\hskip -2pt}c@{\hskip -2pt}c}
 &\ \ \ \ \  Support2 & \ \ \ \ \ MiniBooNE & \ \ \ \ \ News Popularity & \ \ \ \ \ YearPredictionMSD \\
\raisebox{3.7\normalbaselineskip}[0pt][0pt]{\rotatebox[origin=c]{90}{Zero-out}}&
\includegraphics[width=0.24\linewidth]{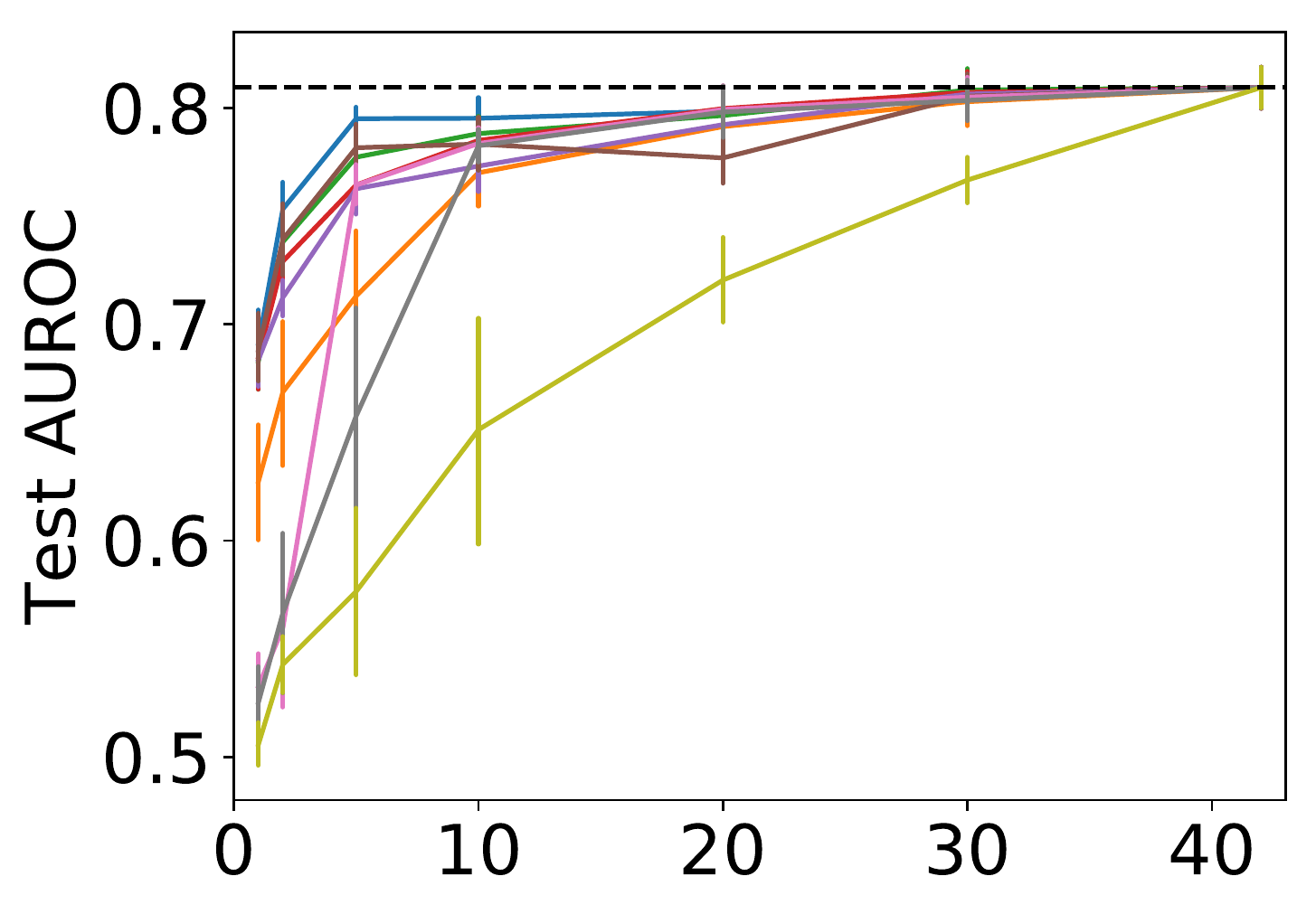}&
\includegraphics[width=0.24\linewidth]{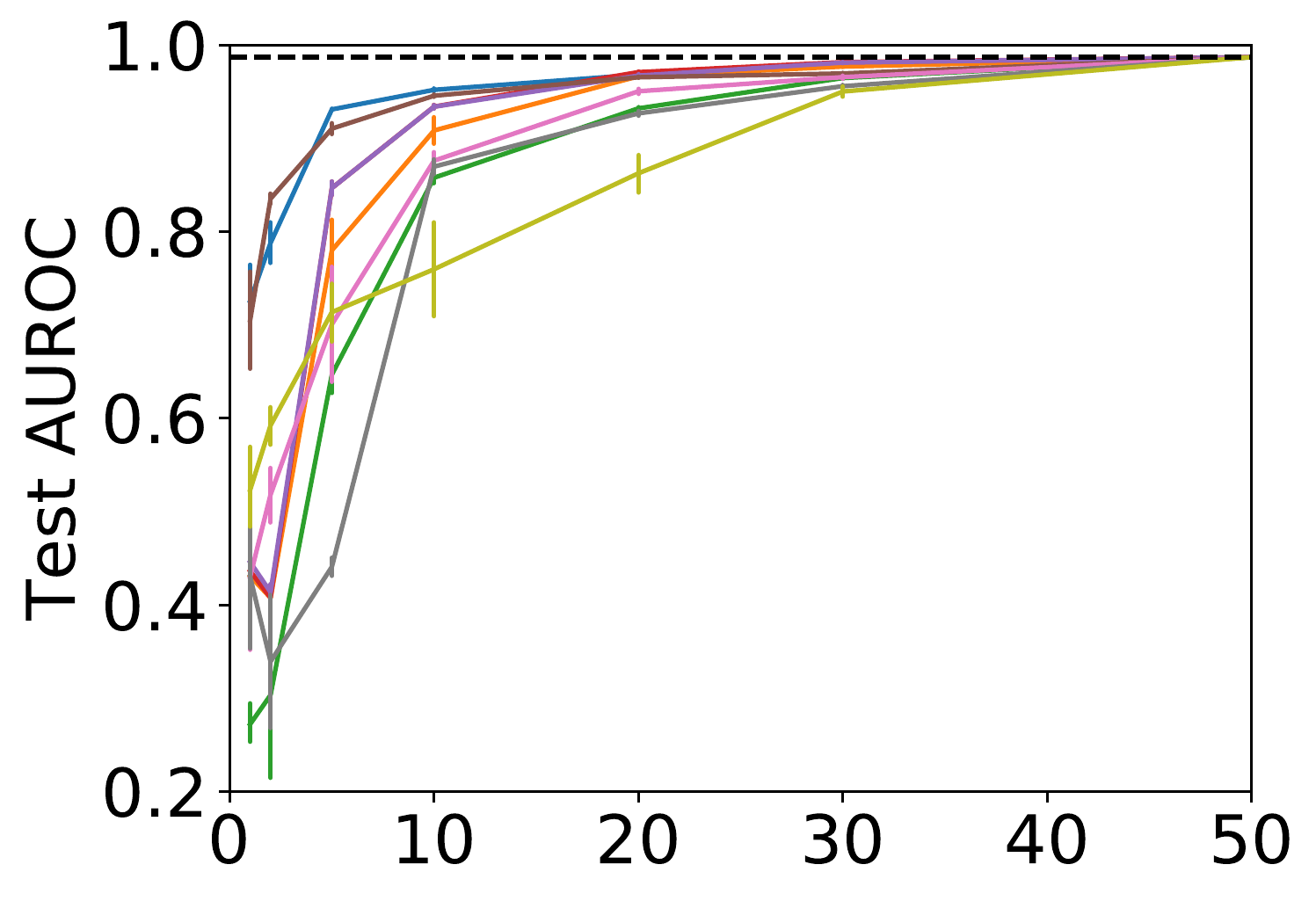}&
\includegraphics[width=0.24\linewidth]{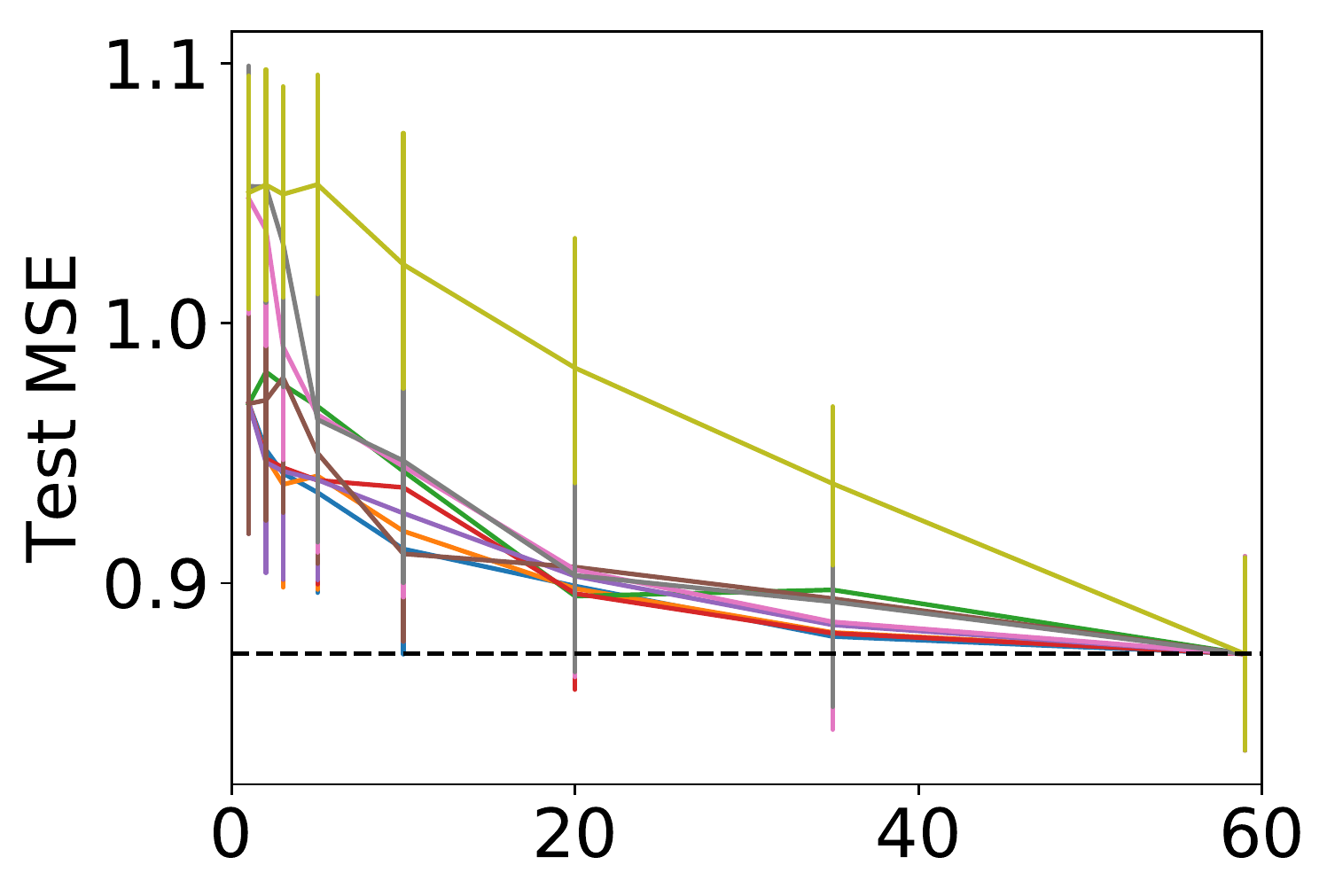}&
\includegraphics[width=0.24\linewidth]{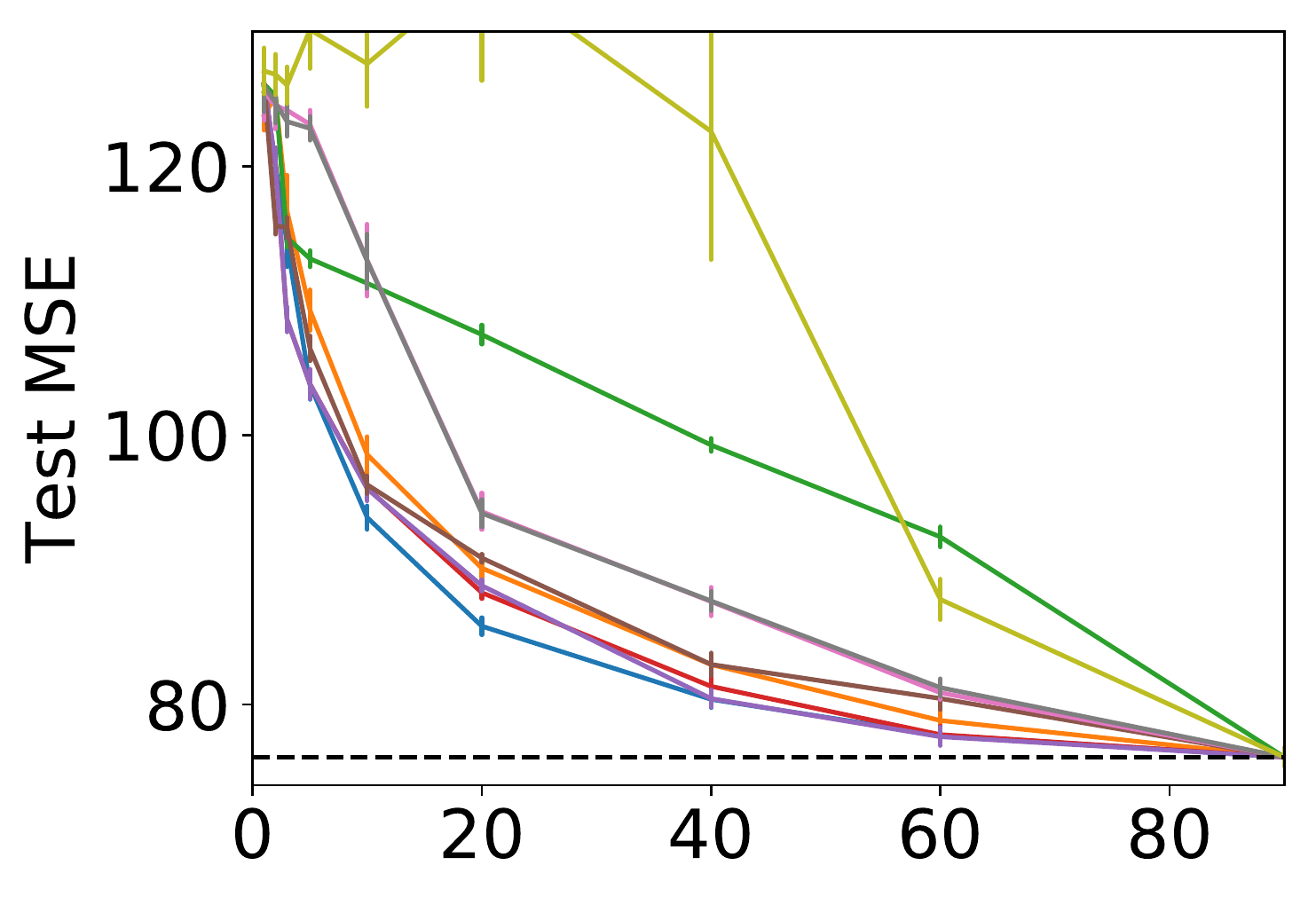}\\
\raisebox{3.7\normalbaselineskip}[0pt][0pt]{\rotatebox[origin=c]{90}{Retrain}}&
\includegraphics[width=0.24\linewidth]{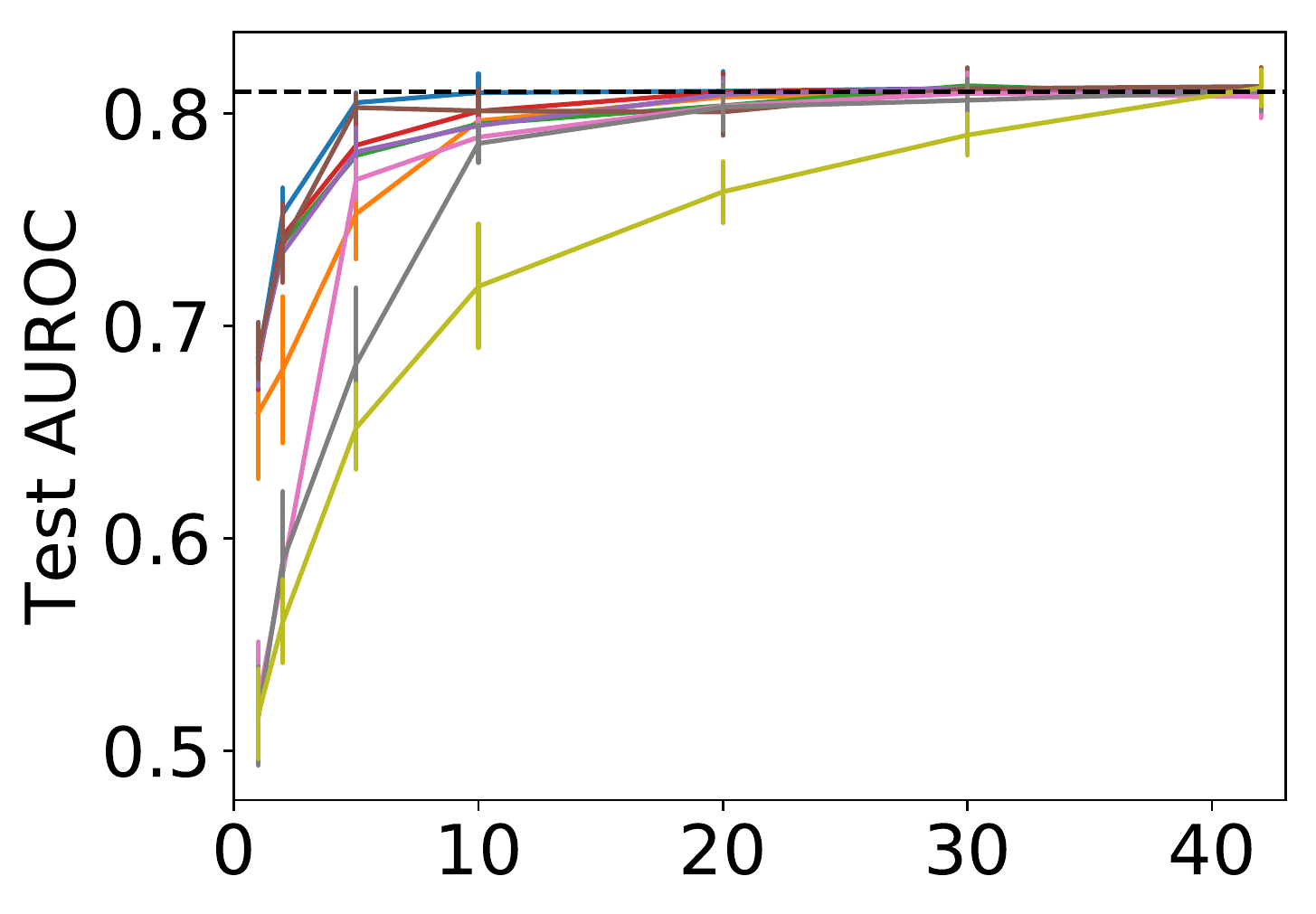}&
\includegraphics[width=0.24\linewidth]{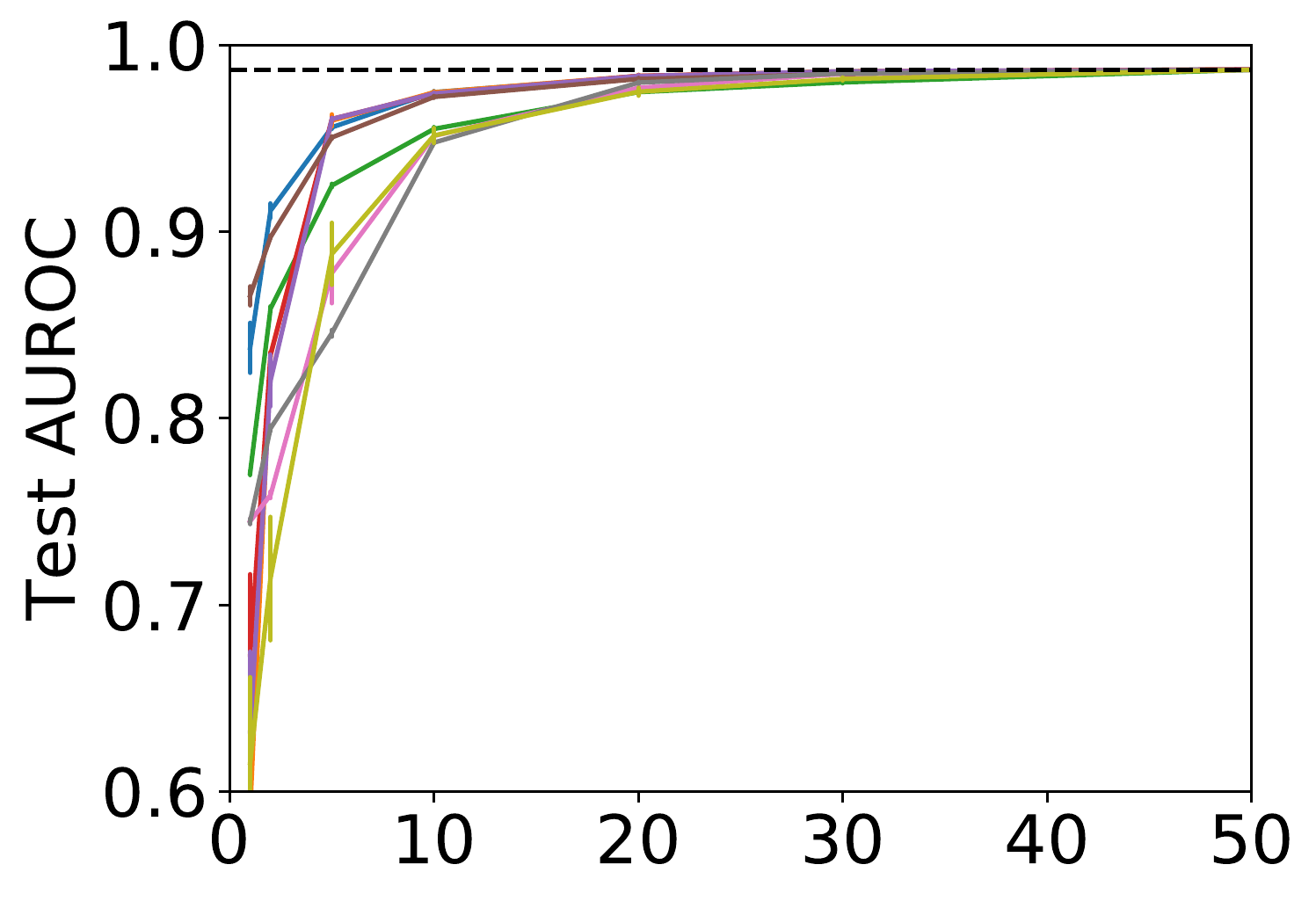}&
\includegraphics[width=0.24\linewidth]{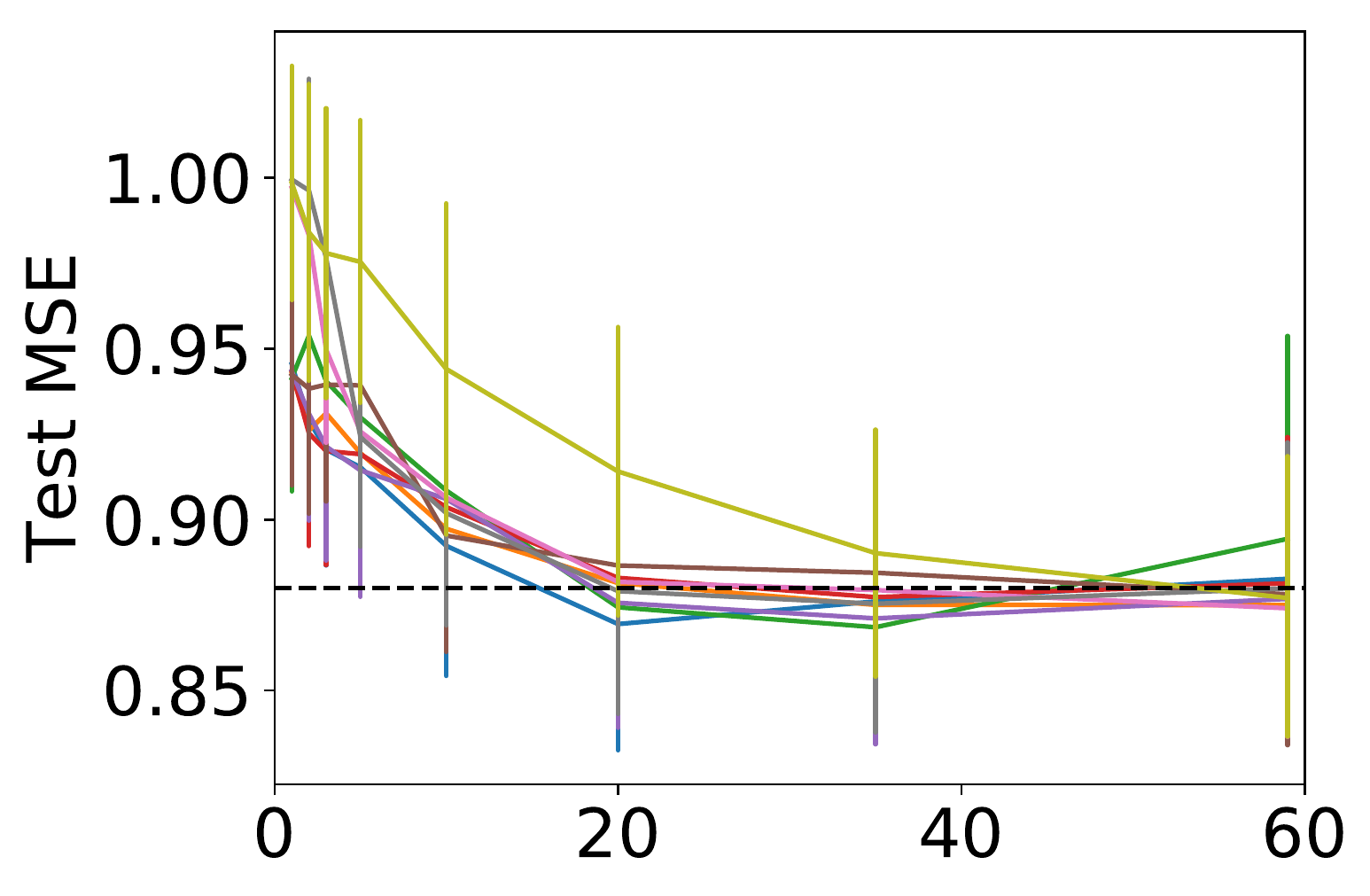}&
\includegraphics[height=80pt,width=0.19\linewidth]{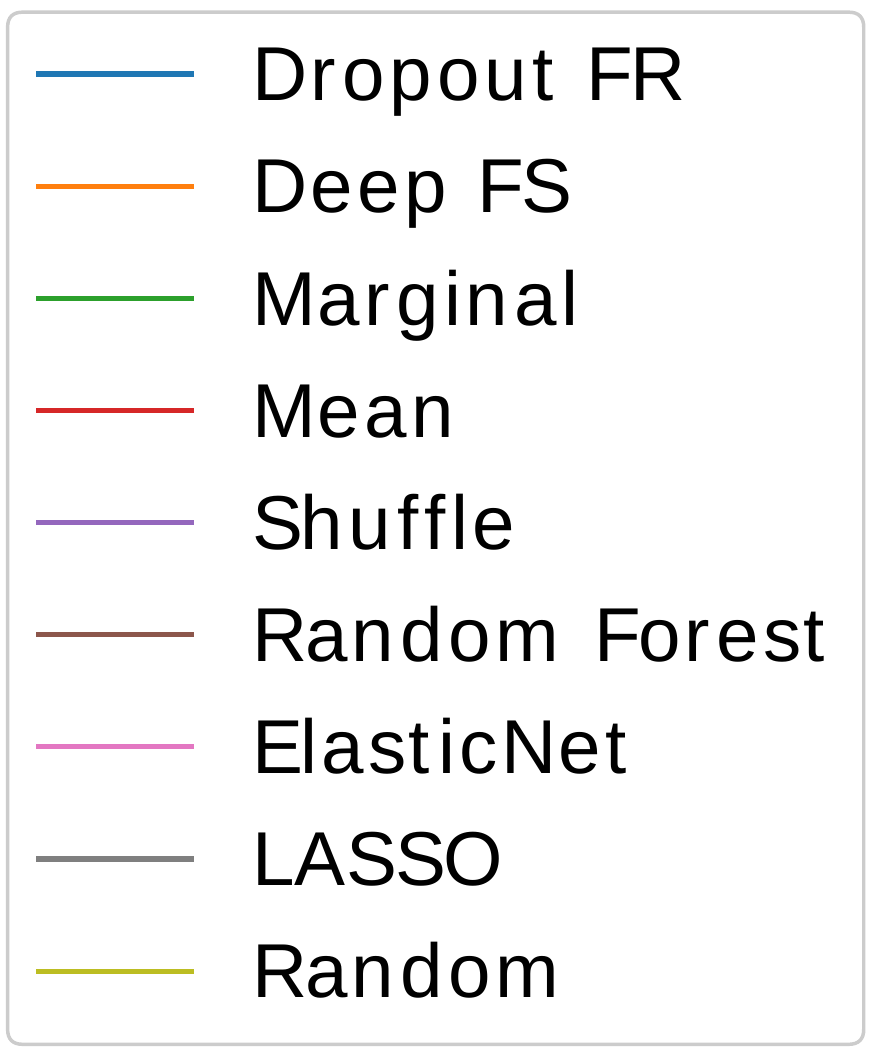}\\
 & \ \ \ \ \ \ \ Number of features & \ \ \ \ \ \ \ Number of features & \ \ \ \ \ \ \ Number of features & 
\end{tabular}
    \caption{Comparison of methods on $4$ datasets with $2$ evaluation settings (zero-out, retrain).}
    \label{fig:uci} 
\end{figure*}

In Figure \ref{fig:uci}, we compare our ranking methods with all other methods mentioned in section \ref{sec:other_methods}. In the `zero-out' setting (first row), our method compares favorably in all the $4$ datasets we tested, with significant difference in the larger dataset YearPredictionMSD.
We note that we get slightly inferior performance to that of random forest when only the top $1$ or $2$ features are used in the MiniBooNE dataset. We also observe similar phenomenon compared to Shuffle method in the YearPredictionMSD dataset. However, the overall performance on these and other datasets, when only the top $1$ or $2$ features are used, is much worse compared to the performance with $5$ or more features, indicating that $1$ or $2$ features are not sufficient to model any of these datasets. Our method has significantly higher performance when the number of features for the same and other dataset is $5$ or higher. 
We deduce that Dropout FR selects better combinations of features (since it gets lower loss as the number of features gets larger) at the cost of the performance just given the top few features.

In the 'retrain' setting (second row), we only compare the first $3$ datasets due to the time it takes to retrain models for the YearPredictionMSD dataset.
In this setting, we find that our method rivals or outperforms other methods.
This demonstrates that Dropout FR method can retrieve better feature combinations suited to the neural network architecture than many other approaches in the wide variety of datasets.

In both settings, we find that marginal ranking (green) performs much better in Support2 and News Popularity dataset and much worse in more complicated datasets, MiniBooNE and YearPredictionMSD. 
It might also be the reason why Dropout FR performs relatively close to other baselines in these $2$ simpler datasets since using marginally important features is sufficient to explain the outcome.
However, as datasets get bigger and more complicated, our method achieves significantly better results than other baselines as seen in MiniBooNE (only RF is close) and YearPredictionMSD dataset. Note that these comparisons also help us to infer the complexity of the datasets, thus it maybe beneficial to gain more insights into the data by always evaluating the performance of the strawmen methods alongside Dropout FR.

\subsubsection{Stability}
\begin{figure}[H]
\begin{center}
\begin{tabular}{Dc}
\raisebox{4.\normalbaselineskip}[0pt][0pt]{\rotatebox[origin=c]{90}{$1 - P_{dropout}$}} &
\includegraphics[width=0.5\linewidth]{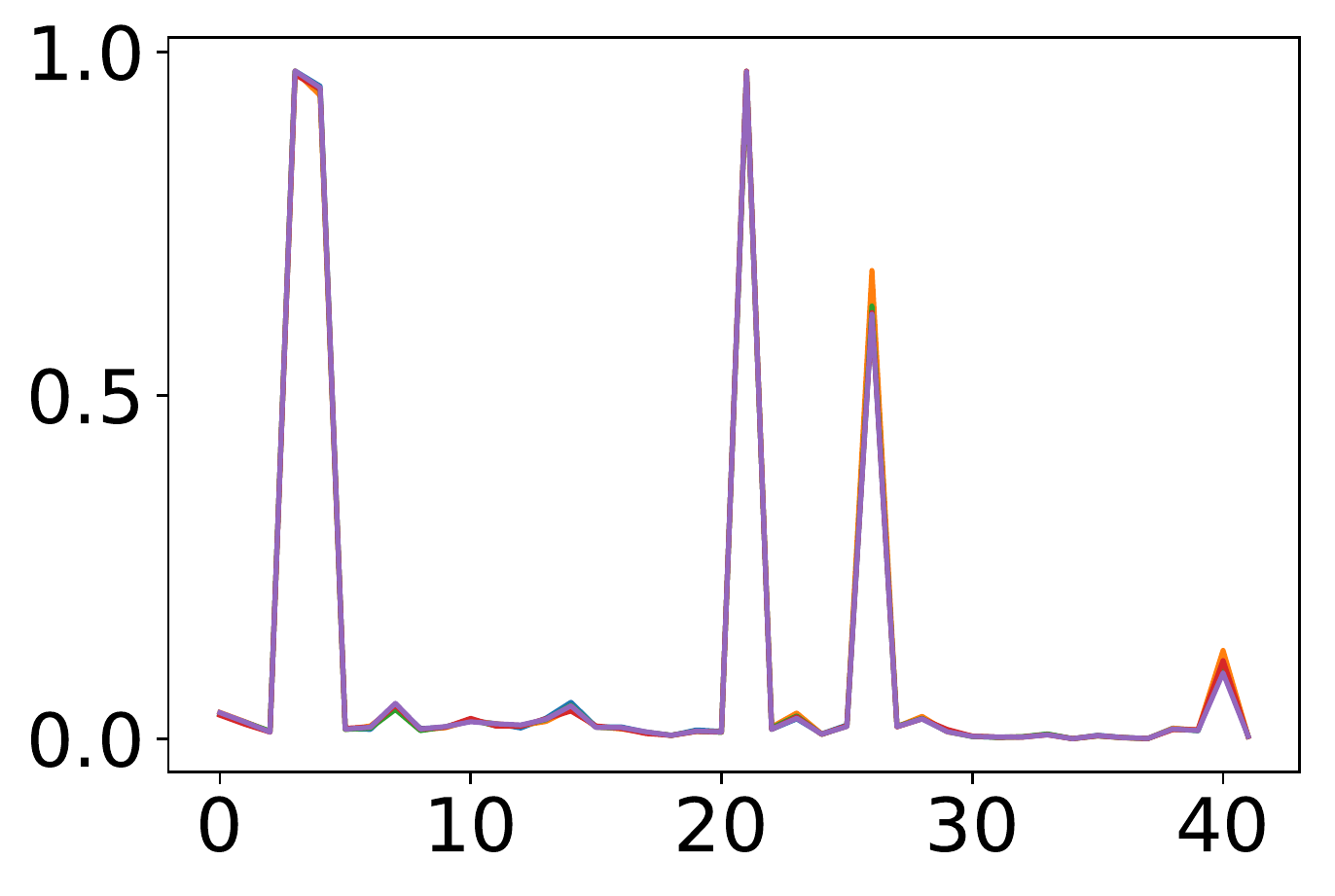} \\
 & \ \ \ \ \ Feature Index \\
\end{tabular}
\vskip -5pt
\caption{The keep probability ($1 - p_{dropout}$) of $47$ features in the Support2 dataset for $5$ different random seeds. ($5$ different colors) when $\lambda = 0.1$.}
\label{fig:random_seed}
\end{center}
\end{figure}

In this experiment, we show that our algorithm is robust to random initialization of the neural network.
Figure \ref{fig:random_seed} shows $5$ different runs with different random seeds in the Support2 dataset setting $\lambda=0.01$. We show that they all converge to similar dropout rate  for each feature after optimization (shown by complete overlap of the performances corresponding to $5$ different seeds on the graph).

\subsubsection{Regularization Coefficient Effect}
\begin{figure}[H]
\begin{center}
\begin{tabular}{Dc}
\raisebox{4.7\normalbaselineskip}[0pt][0pt]{\rotatebox[origin=c]{90}{$1 - P_{dropout}$}} &
\includegraphics[width=0.45\textwidth]{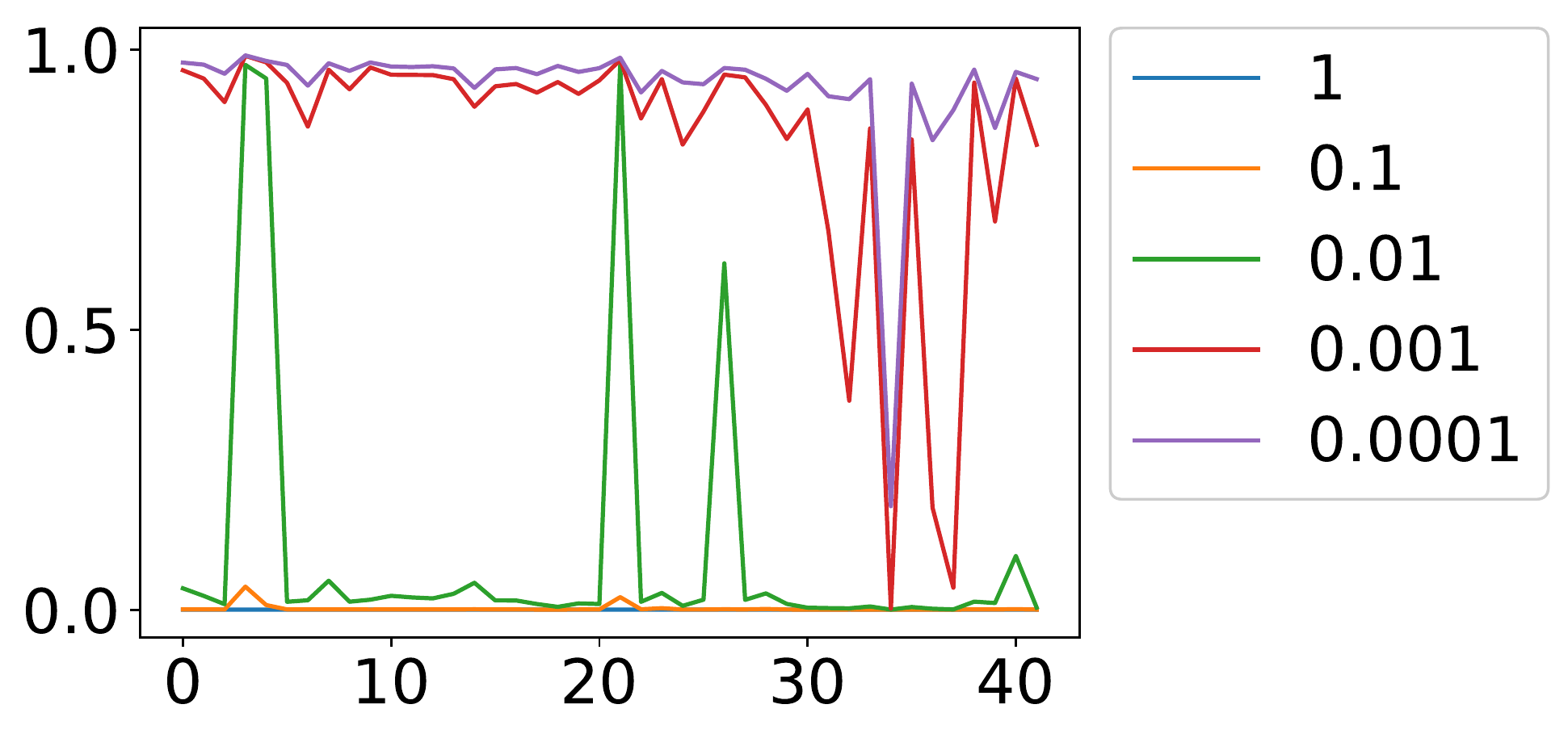} \\
 & \hspace{-50pt} Number of Features \\
\end{tabular}
\caption{The keep probability of $47$ features in the Support2 dataset for different regularization coefficient $\lambda$.}
\label{fig:reg_coef}
\end{center}

\end{figure}

In Figure \ref{fig:reg_coef}, we examine the effect of different regularization coefficient on the final dropout rate in our algorithm.
We note that when we have strong regularization (high $\lambda$), most of the features get pruned and have high dropout rate (low keep probability). On the other hand, when the regularization is too weak, every instance has the dropout rate close to $0$.
It is crucial to select proper $\lambda$ that preserves the important features while pruning the noisy features.

\subsection{Predicting in-hospital mortality}
\label{sec:hospital}
In this experiment, we evaluate the performance of our method using a multivariate time-series clinical dataset to determine the importance of clinical covariates in predicting in-hospital mortality. 
This dataset, from PhysioNet 2012 Challenge \citep{goldberger2000physiobank}, is a publicly available collection of multivariate clinical time series with $8000$ intensive care unit (ICU) patients. 
It contains $37$ patient measurements within the first $48$ hours in the ICU. The goal is to predict the in-hospital mortality as a binary classification problem. 
We use the only publicly available \emph{Training Set A} subset which contains $4,000$ patient measurements with $554$ patients having the positive mortality labels.

\newcolumntype{E}{ >{\centering\arraybackslash} m{0.1cm} }

\begin{figure*}[tbp]
   \centering
\begin{tabular}{Ec@{\hskip -3pt}c}
 &\ \ \ \ \  \large Physionet (AUROC) & \ \ \ \ \ \large Physionet (AUPR)  \\
\raisebox{3.3\normalbaselineskip}[0pt][0pt]{\rotatebox[origin=c]{90}{\large Zero-out}}&
\includegraphics[width=0.3\textwidth]{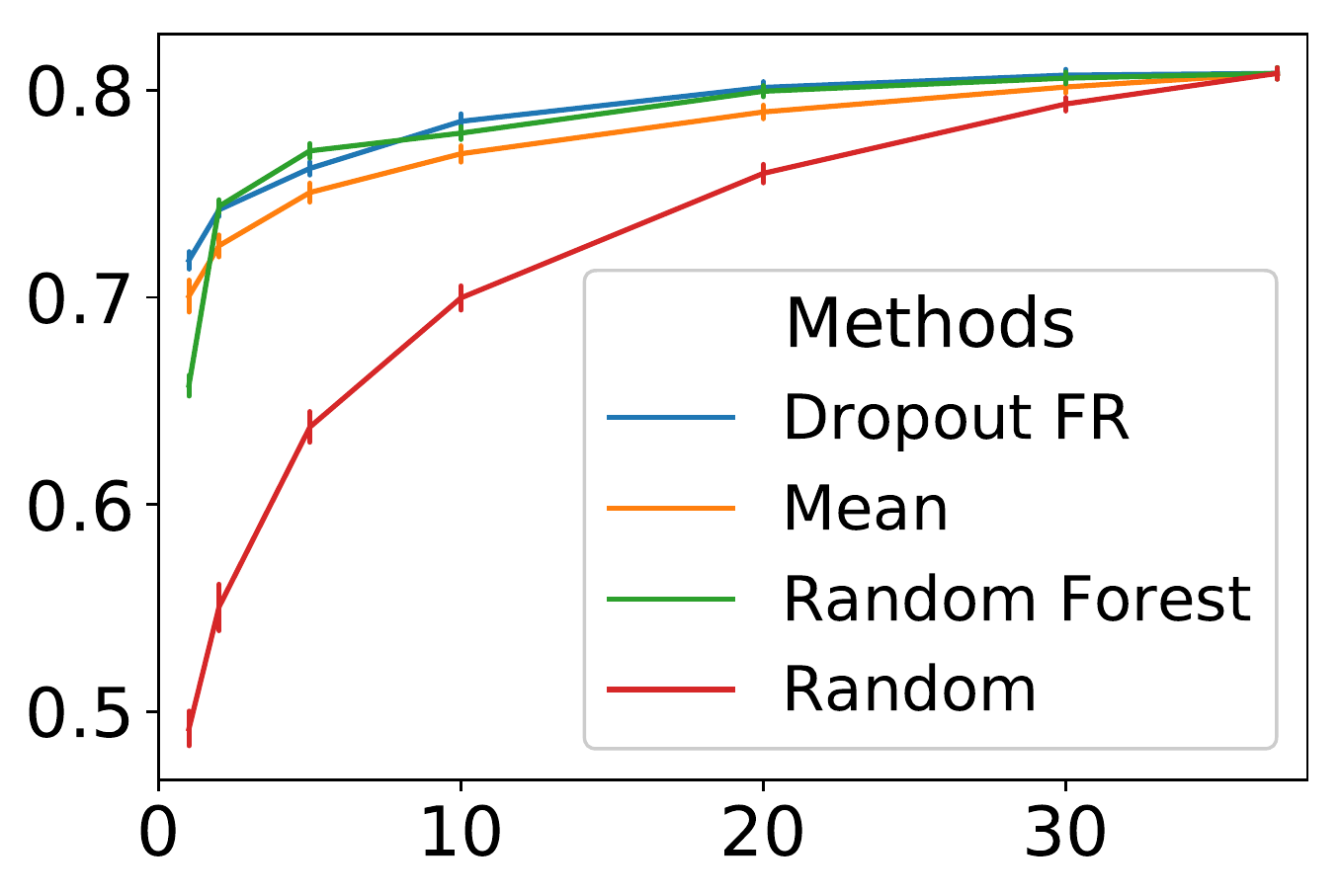}&
\includegraphics[width=0.3\textwidth]{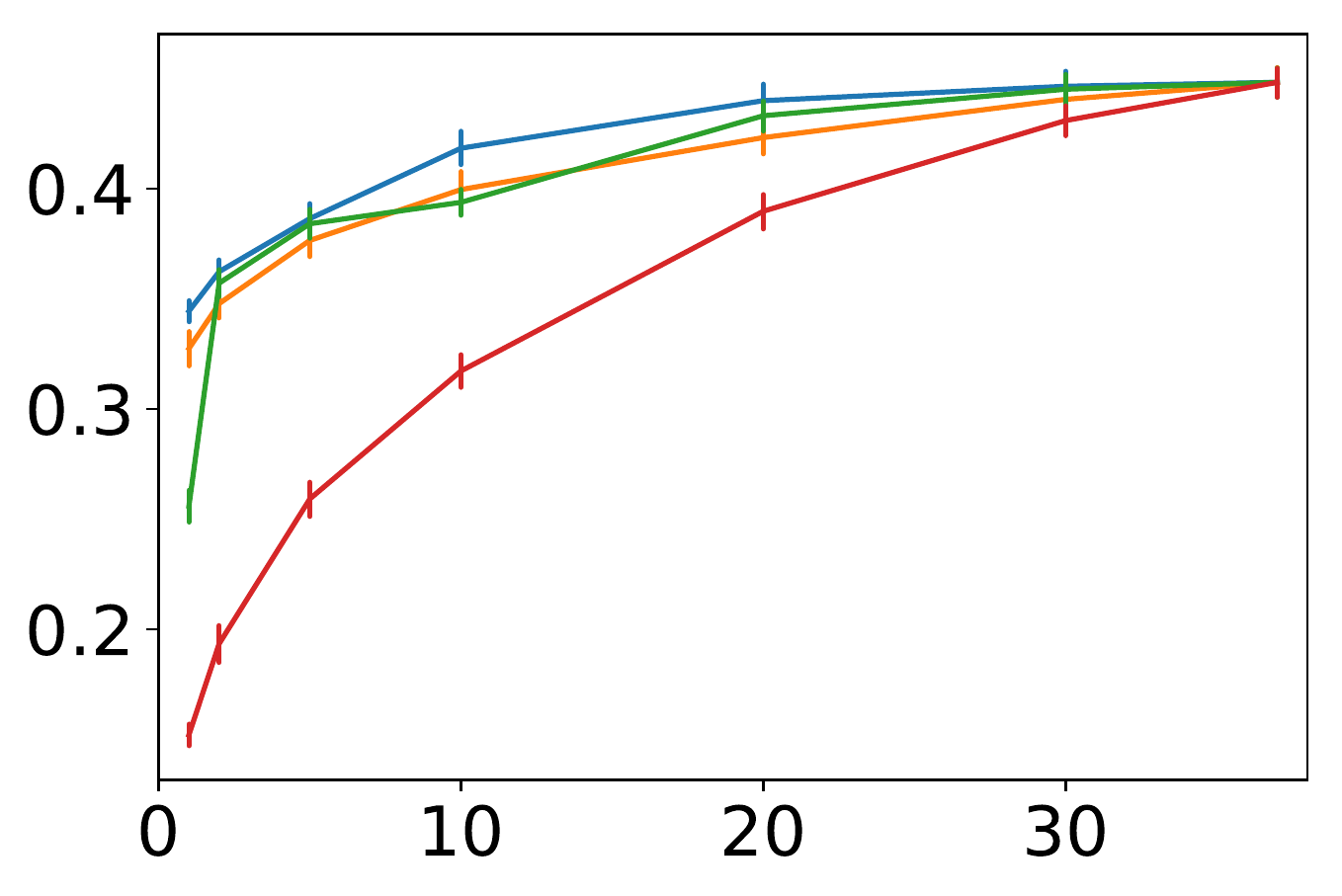}\\
\raisebox{3.\normalbaselineskip}[0pt][0pt]{\rotatebox[origin=c]{90}{\large Retrain}}&
\includegraphics[width=0.3\textwidth]{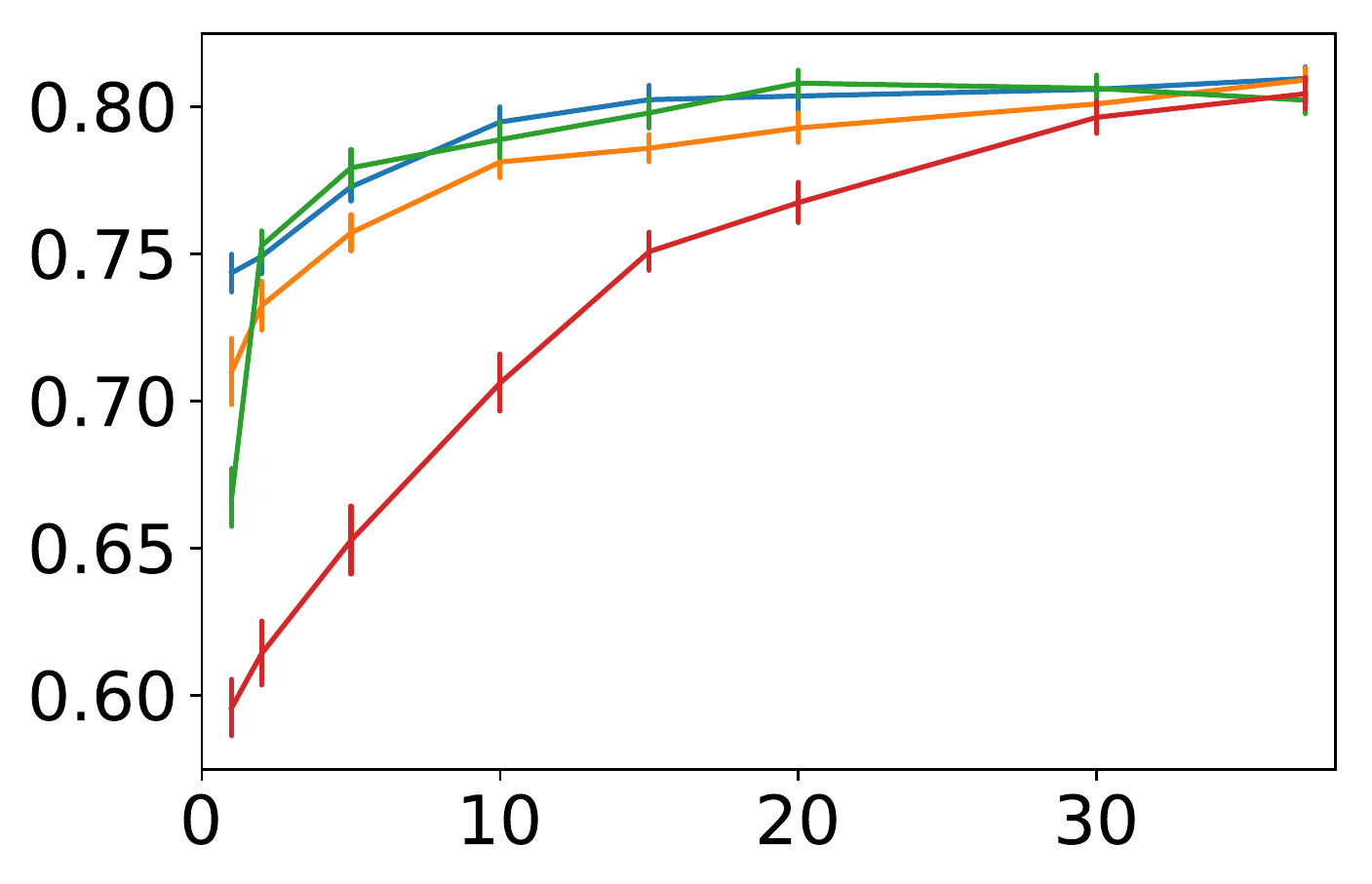}&
\includegraphics[width=0.3\textwidth]{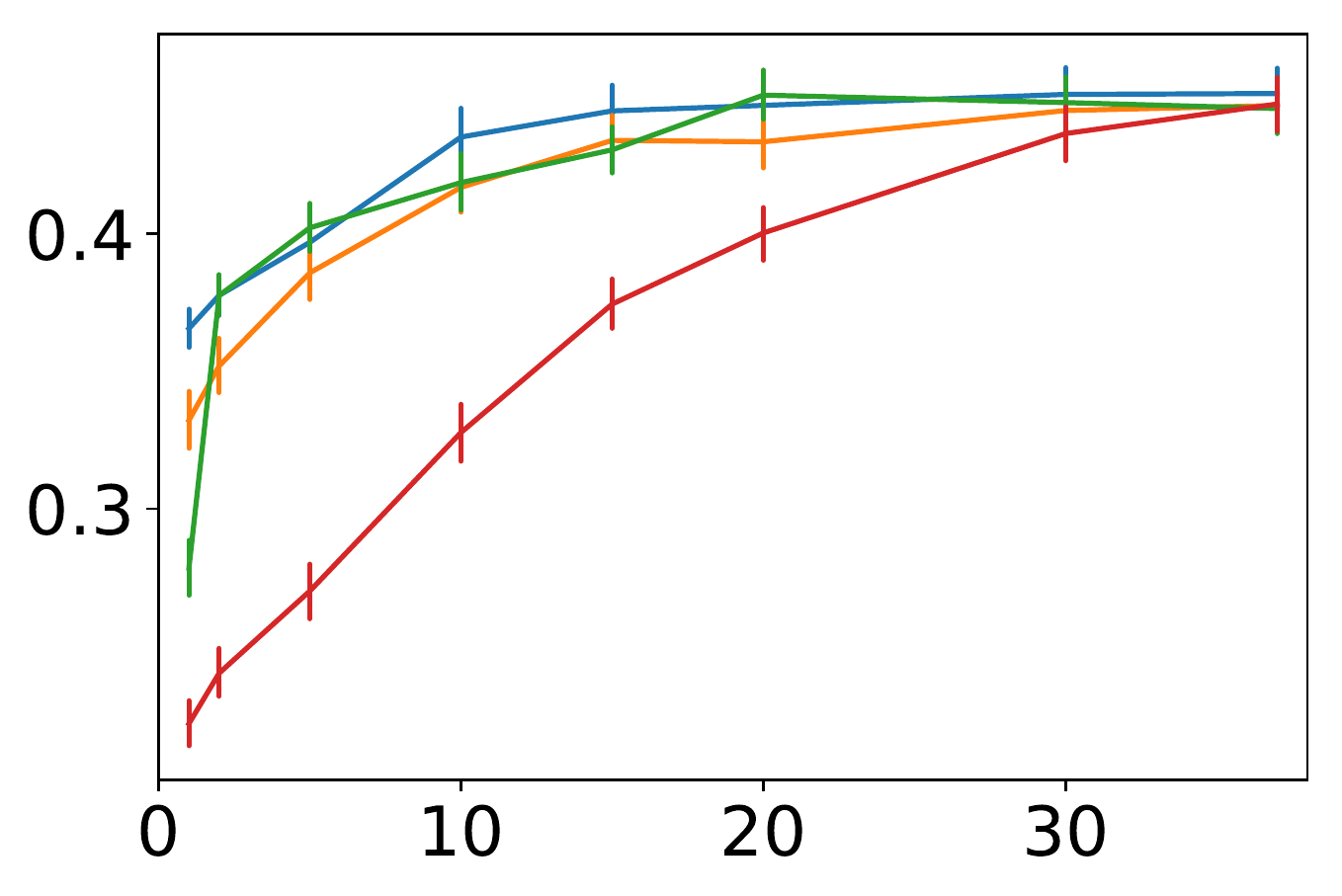}\\
 &\ \ \ \ \ \ \  \large Number of features & \ \ \ \ \ \ \ \large Number of features
\end{tabular}
    \caption{Comparison of methods on Physionet datasets with both AUROC and AUPR.}
    \label{fig:physionet} 
\end{figure*}

We follow the preprocessing of \citet{lipton2016modeling} work. First, we use binary features indicating whether or not a feature was measured at a given time point. If a feature was not measured, we set the binary variable to $1$ and if it was measured, we set it to $0$. 
Concatinating these reverse-indicator variables with the original $37$ features results in $74$ features in total. 
Second, we normalize each feature to zero mean and unit variance except for the binary features.
Finally, we bin the input features into 1-hour intervals, take average of multiple measurements within 1-hour time window, and impute missing values with $0$. 
These lead to a time-series with $48$ time points and $74$ features.
We split the dataset randomly into $80\%$, $10\%$, and $10\%$ as training, validation and test set, respectively, and repeat the procedure $40$ times.

We follow the RNN architecture used in \citet{che2016recurrent} to predict the mortality 
.
We use $10$-fold cross validation to select $\lambda$.
For random forest, we use $1,000$ trees and sum the feature importance across all the time points (since in RF each time point is considered independently for each of the features), including original feature and its corresponding reverse-indicator features.

First, we compare the neural network performance with respect to other commonly-used classifiers on the PhysioNet dataset and show that RNN is better in test set AUPR and AUROC in Table \ref{table:performance}.


In Figure \ref{fig:physionet}, we compare Dropout FR, Mean, RF ranking method and Random ranking in the zero-out and retrain settings with both AUROC and AUPR.
We find that Dropout FR performs overall better than Random Forest with significant difference in using $1$ feature across all settings we evaluate.
We also find that Dropouot FR performs significantly better than Mean method across all settings.
Overall, we show that our method performs well in the recurrent neural network architecture, capturing the feature importances in the time-series datasets.

In Table \ref{table:physionet_top_features}, we show the top $10$ features selected by RF and Dropout FR. 
Overall these two approaches rank features somewhat differently, though many of the features in the two lists are the same. 
We find the reason for the inferior RF performance observed in Figure \ref{fig:physionet} when only $1$ feature is used is the different ranking of `Urine' and `GCS' features (RF selects `GCS' as second).
The table also demonstrates that feature importance does not simply follow the frequency of the features in the dataset for either of the methods.


\vspace{-5pt}
\subsection{Drug Response Prediction}
We apply our method to a real-world drug response dataset to find which genes determine drug response using the semi-supervised variational autoencoder (SSVAE) \citep{kingma2014SSVAE} model applied to this task by \citet{rampasek_drvae_2017}, who kindly shares the dataset and code with us.
The SSVAE takes gene expression of 903 preselected genes as input and performs a binary classification to find whether the given cell line responds to the drug. 

In Table \ref{table:drvae}, we examined genes contributing to the response of bortezomib, a drug commonly used in multiple myeloma patients.
We choose this drug since the model performs the best in this drug and it is widely investigated in biological research literature.
The gene that was ranked the highest by our algorithm (with lowest dropout probability), \textit{NR1H2}, was previously found to be indicative of Multiple Myeloma (MM) non-response to anti- agents such as bortezomib \citep{agarwal2014activation}. The second ranked gene, \textit{BLVRA}, is known to be amplified in cells sensitive to anti-MM treatment, such as bortezomib \citep{soriano2016proteasome}. Interestingly, \textit{BVLRA} was also ranked second by RF (and not ranked highly by t-test). The gene ranked first by RF is \textit{FOSL1} which was not directly found to be linked to response by bortezomib, but is tangentially related through osteoclass process 
(\textit{FOSL1} helps with differentiation into bone cells and there is a secondary effect of bortezomib to prevent bone loss during inflammation processes). Overall, we found that ranking of RF follows rather closely ranking by t-test. Dropout FR ranking was significantly different, capturing the importance of the ranking for the SSVAE classification.

\begin{table}[tb]
\processtable{Test set performance of different methods in the Physionet\label{table:performance}} {
\begin{tabular}{lll}
    \toprule
    \cmidrule{1-2}
    Method     & AUPR     & AUROC \\
    \midrule
    SVM-linear & $0.317\pm0.063$  & $0.709\pm0.030$     \\
    SVM-RBF     & $0.428\pm0.070$ & $0.790\pm0.024$  \\
    RF     & $0.415\pm0.062$	& $0.796\pm0.023$ \\
    RNN     & $\bm{0.448}\pm\bm{0.063}$	& $\bm{0.808}\pm\bm{0.026}$ \\
    \bottomrule
  \end{tabular}
}{}
\vspace{-20pt}
\end{table}

\vspace{-5pt}
\section{Discussion}
In this work we proposed a new general approach for understanding the importance of features in deep learning. This simple approach has been previously shown to be very powerful for regularizing DNNs and preventing them from overfitting, but thus far has never been used on the input layer or applied to the task of feature ranking, i.e. to understand the performance of DNNs. We believe that variational dropout works well because it acts similarly to feature bagging \citep{1995_random_bagging}, subsetting the features during training. It allows to decouple correlated variables in certain instances and optimizes the corresponding feature-wise dropout rate. This may also be the reason for the good performance by random forest which we have observed in our experiments and also the reason for poor performance of $\ell_1$ used in LASSO and Deep FS. 

In our simulation experiment, we showed that deep learning based methods (Dropout FR, Zero, and Shuffle) capture the second-order interactions well.
For other methods, we find that Random Forest performs worse when considering the order of more important features, showing it is not able to capture the correct ranking among important interacting features.
Other methods such as Marginal, LASSO and Elastic Net perform much worse in our simulation, indicating simple univariate testing or linear layer is not sufficient to capture complicated nonlinear effects across both simulated and real datasets.

Further, we tested our approach in $4$ feed-forward networks, a recurrent neural network and a semi-supervised variational autoencoder showing that Dropout FR is applicable to various deep learning architectures and in most scenarios performs better than other commonly-used baselines and in other scenarios it performs as well as some of the best alternatives.
In particular, our experiments in the feed-forward neural networks show that our method outperform other methods significantly in the MiniBooNE (ranking the top $5$ features) and YearPredictionMSD (ranking top $20$ features) datasets.
Although we find our approach is not the best performer for some small numbers of features, we consider it reasonable since it is not a greedy approach and thus might optimize the ranking that sacrifices the performance of fewer features in exchange for larger performance gain for a bigger combination of features.
In addition, in the time-series setting (Physionet), our approach outperforms other methods, including Random Forest when only using the top one feature. We see the same phenomenon in simulation that Random Forest is not good at ranking the top-ranked features, which is important for experimental design. Overall, we found it useful to compare multiple strawmen, such as marginal ranking, to gain further insights into the complexity of the data.

\begin{table}[t]
\processtable{Top $10$ features selected from RF rank and Dropout FR in Physionet\label{table:physionet_top_features}} {
\begin{tabular}{lllll}
    \toprule
 & RF rank   & Present & RNN rank  & Present \\
    \midrule 
1    & Urine     & $1.87\%$              & GCS       & $0.86\%$              \\
2    & GCS       & $0.86\%$              & Urine     & $1.87\%$              \\
3    & HR        & $2.43\%$              & BUN       & $0.20\%$              \\
4    & SysABP    & $1.46\%$              & MechVent  & $0.41\%$              \\
5    & Temp      & $1.00\%$              & Temp      & $1.00\%$              \\
6    & NISysABP  & $1.14\%$              & HR        & $2.43\%$              \\
7    & NIMAP     & $1.12\%$              & Lactate   & $0.11\%$              \\
8    & Weight    & $1.43\%$              & Weight    & $1.43\%$              \\
9    & NIDiasABP & $1.14\%$              & NIDiasABP & $1.14\%$              \\
10   & MAP       & $1.45\%$              & SysABP    & $1.46\%$             \\
 \bottomrule
\end{tabular}
}{}
\vspace{-20pt}
\end{table}

\begin{table}[t]
\processtable{Top $10$ genes selected from RF rank and Dropout FR in drug response. We also list the t-test pvalue for each gene.\label{table:drvae}} {
\begin{tabular}{lllll}
    \toprule
   & RF rank  & Pvalue   & NN rank  & Pvalue   \\
    \midrule 
1  & \textit{FOSL1}    & 1.48e-07  & \textit{NR1H2}    & 2.26E-05 \\
2  & \textit{BLVRA}    & 1.30e-01  & \textit{BLVRA}    & 1.30E-01 \\
3  & \textit{TRAM2}    & 1.18e-10  & \textit{PIK3CA}   & 6.14E-02 \\
4  & \textit{CD44}     & 1.21e-07  & \textit{ATP6V1D}  & 4.34E-01 \\
5  & \textit{DRAP1}    & 1.30e-06  & \textit{USP20}    & 8.47E-02 \\
6  & \textit{FKBP4}    & 4.99e-04  & \textit{TRAM2}    & 1.18E-10 \\
7  & \textit{UBE2L6}   & 2.77e-06  & \textit{CD58}     & 2.20E-05 \\
8  & \textit{PLEKHM1}  & 8.83e-04  & \textit{DHX8}     & 3.95E-02 \\
9  & \textit{SERPINE1} & 8.34e-05  & \textit{PPP1R13B} & 5.40E-03 \\
10 & \textit{BAX}      & 3.05e-05 & \textit{CASP10}   & 6.07E-03 \\
 \bottomrule
\end{tabular}
}{}
\vspace{-20pt}
\end{table}

\section{Conclusion}
We propose a new general feature ranking method for deep learning to interpret the feature importance. When it is impossible to measure all the features under various constraints, such as limited time and undue physical or emotional burden on the patient, it is paramount to design the system that collects the right subsets of features leading to highest performance. 
Our method can be used to design diagnostic standard procedure that measures least number of clinical tests with the highest or comparable predictive power.
In conclusion, we provide a new and effective method that addresses the resource-constraint setting which is widely seen in the healthcare industry, and an effective solution to the common need in biology to interpret the predictive system especially such as deep learning, commonly thought of as a complex black box.










\bibliography{ref}

\bibliographystyle{natbib}

\end{document}